\documentclass{article}

\usepackage[final]{d3s3_neurips_2024}

\usepackage[utf8]{inputenc} 
\usepackage[T1]{fontenc}    
\usepackage{hyperref}       
\usepackage{enumitem}
\usepackage{url}            
\usepackage{booktabs}       
\usepackage{nicefrac}       
\usepackage{microtype}      
\usepackage{xcolor}         
\usepackage{natbib}
\usepackage{tabularx,ragged2e,booktabs,caption}
\usepackage{subfigure}
\usepackage{multicol,tabularx,capt-of}
\usepackage{hhline}
\usepackage{wrapfig}
\usepackage{multirow}
\usepackage{multicol}
\usepackage{bm}
\usepackage{nomencl} 
\usepackage{algorithm}
\usepackage{algorithmic}
\usepackage{subcaption}
\usepackage{mathtools}
\usepackage{setspace}
\usepackage{natbib}

\title{Using Parametric PINNs for Predicting Internal and External Turbulent Flows}

\author{%
  Shinjan Ghosh$,$ Amit Chakraborty$,$ Georgia Olympia Brikis$,$ Biswadip Dey \\
  Siemens Corporation\\
  Princeton, NJ 08540, USA. \\
}

\begin{document}

\maketitle

\begin{abstract}
Computational fluid dynamics (CFD) solvers employing two-equation eddy viscosity models are the industry standard for simulating turbulent flows using the Reynolds-averaged Navier-Stokes (RANS) formulation. While these methods are computationally less expensive than direct numerical simulations, they can still incur significant computational costs to achieve the desired accuracy. In this context, physics-informed neural networks (PINNs) offer a promising approach for developing parametric surrogate models that leverage both existing, but limited CFD solutions and the governing differential equations to predict simulation outcomes in a computationally efficient, differentiable, and near real-time manner. In this work, we build upon the previously proposed RANS-PINN framework, which only focused on predicting flow over a cylinder. To investigate the efficacy of RANS-PINN as a viable approach to building parametric surrogate models, we investigate its accuracy in predicting relevant turbulent flow variables for both internal and external flows. To ensure training convergence with a more complex loss function, we adopt a novel sampling approach that exploits the domain geometry to ensure a proper balance among the contributions from various regions within the solution domain. The effectiveness of this framework is then demonstrated for two scenarios that represent a broad class of internal and external flow problems.
\end{abstract}

\section{Introduction}

In recent years, deep learning has become an effective means for fast prediction of the outcome of CFD simulations \citep{vinuesa2022enhancing, warey2020data, zhang2022demystifying}. While some approaches leverage deep learning to enhance traditional CFD solvers \citep{hsieh2019learning, doi:10.1073/pnas.2101784118}, others learn to predict flow fields using convolutional or graph neural networks on grids or meshes \citep{hsieh2019learning, hennigh2017lat, jiang2020meshfreeflownet, wang2020towards}. On the other hand, physics-informed neural networks (PINNs) embed governing PDEs into a regularization term while learning to predict flow variables over the spatiotemporal continuum \citep{raissi2019physics, dwivedi2019distributed, lu2019deepxde, nabian2019deep, zhang2020frequencycompensated}.

While these advancements demonstrate the potential of deep learning in fluid dynamics, the accurate prediction of turbulent flow, particularly in RANS-based turbulence modeling, remains a critical challenge that demands further exploration. However, comprehensive studies on employing PINNs in the context of RANS-based turbulence modeling, particularly for predicting flow fields for unforeseen Reynolds numbers, remain limited \citep{en16041755}. While using PINNs to predict turbulent flow fields over a solution domain, previous studies have used the Reynolds-stress formulation, which relies on high-quality data from computationally expensive simulations such as Direct Numerical Simulations, Large Eddy Simulations, or high-resolution flow measurements \citep{eivazi2022physics, HANRAHAN}. On the other hand, \citet{patil} explored the assimilation of data and physics within the one-equation Spalart-Allmaras (SA) turbulence model. \citet{pioch} investigated the efficacy of multiple turbulence models, including $k$-$\omega$, Reynolds-stress, and model-free cases, by employing both data-free and DNS data based hybrid training to reconstruct flow fields for a single Reynolds number. Most recently, \cite{ghosh2023_RANSpinn} used the $k$-$\epsilon$ formulation of the two-equation eddy viscosity model and explored its use for flow reconstruction and also to predict turbulent flow for unforeseen Reynolds numbers for the cylinder in a cross-flow problem. 

In this work, we adopt a similar approach and investigate the effectiveness of the $k$-$\epsilon$ turbulence model in building a parametric surrogate model that can predict turbulent flow fields across unforeseen Reynolds numbers for both internal and external flow geometries. Given the complexity of flow prediction problems, relying solely on a single scalar quantity is insufficient to evaluate the learned surrogate models comprehensively. Therefore, we leverage macroscopic analyses of flow features, including stagnation zones, wake turbulence, and separation bubbles, to gain deeper insights. In particular, we use data from a limited set of RANS simulations to build a parametric surrogate model and assess its accuracy in predicting flows both within and beyond the Reynolds number range used in its training. The main contributions of this work are as follows:
\begin{itemize}[noitemsep,topsep=-0.5em]
\item Incorporation of $k$-$\epsilon$ turbulence model into a PINN-based framework for building parametric surrogates of turbulent flows.
\item Implementation of a carefully designed warm-start phase to ensure convergence when dealing with multiple PDE-based loss terms along with limited CFD data.
\item Accurate prediction of velocity and pressure fields for a given Reynolds number (both within and beyond the training range) over multiple flow geometries.
\item Development of a systematic approach to analyze the spatial distribution of prediction error and subsequently understand the quality of prediction (inside and outside training range) within the solution domain. 
\end{itemize}

\section{RANS-PINN Architecture and Training}
\label{gen_inst}

The flow of an incompressible fluid is governed by the continuity and Navier-Stokes equations, which enforce the conservation of mass and momentum, respectively. Furthermore, to capture the effect of turbulence, we employ a $k$-$\epsilon$ turbulence model, a specific formulation of the 2-equation eddy viscosity model, to describe the evolution of the closure terms associated with the RANS formulation.

By letting $u_{inlet}$ and $L$ denote the inlet velocity and the characteristic length, respectively, the nondimensional \textit{Reynolds number} for this system can be defined as $\mathcal{R}_e = \frac{u_{inlet} L}{\mu}$. Following this nondimensionalization, we normalize the spatial coordinates, the velocity, and the pressure with the characteristic length, the inlet velocity, and the dynamic pressure, respectively to improve training convergence. The normalized variables can be expressed as
\begin{align} 
\tilde{x} = \frac{x}{L}
,\quad
\tilde{y} = \frac{y}{L}
,\quad
\tilde{U} = \frac{U}{u_{Inlet}}
,\quad
\tilde{p} = \frac{p} {2 \rho u_{Inlet}^2}
,\quad
\tilde{k} = \frac{k} {2 \rho u_{Inlet}^2}
,\quad \textrm{and} \quad
\tilde{\epsilon} = \frac{\epsilon L} {u_{Inlet}^3},
\label{NonDimVars}
\end{align}
where $(\tilde{x},\tilde{y})$, $\tilde{U}=(\tilde{u},\tilde{v})$, $\tilde{p}$, $\tilde{k}$, and $\tilde{\epsilon}$ represent the normalized values of the spatial coordinates, flow velocity, pressure, turbulent kinetic energy, and dissipation rate, respectively. With these normalized variables, the governing PDEs can be represented as the following set of nondimensional equations:
\begin{align} 
\textbf{Continuity:}\quad &
\nabla(\tilde U) 
= 0,
\\
\textbf{Navier-Stokes:}\quad &
(\tilde U\cdot\nabla)\tilde U + \nabla(\tilde p)
-  \frac{1}{\mathcal{R}_e}\nabla^{2} \tilde U 
= 0,
\\
\textbf{2-equation Model ($\tilde{k}$):}\quad &
\nabla ( \tilde U \tilde k)
- \nabla\left[\left(\frac{1}{\mathcal{R}_e}+\frac{\mu_{t}}{u_{inlet}L{\sigma_{k}}} \right)\nabla \tilde k\right] 
- \tilde P_{k} 
+ \tilde \epsilon
= 0, \quad \textrm{and }
\\
\textbf{2-equation Model ($\tilde{\epsilon}$):}\quad &
\nabla ( \tilde U \tilde \epsilon)
- \nabla\left[\left(\frac{1}{\mathcal{R}_e }+\frac{\mu_{t}}{u_{inlet}L\sigma_{\epsilon}}\right)\nabla\tilde \epsilon\right] 
- \left(C_{1}\tilde{P}_{k}+C_{2}\tilde \epsilon\right)\frac{\tilde \epsilon}{\tilde k}
= 0,
\end{align}
where, $C_1 = 1.44$, $C_2 = 1.92$, $\sigma_k = 1$, and $\sigma_{\epsilon} = 1.3$ are empirical model constants and $P_k$ is a production term that depends on the spatial derivates of flow velocity.

The RANS-PINN architecture uses five individual neural networks to predict the flow variables, which can be implemented using multilayer perceptrons (MLPs) \citep{raissi2019physics}, Fourier neural operators (FNOs) \citep{li2021fourier}, or even Kolmogorov-Arnold networks (KANs) \citep{shukla2024comprehensive}. Our experiments used FNOs and MLPs for the airfoil and the backward-facing step geometry, respectively, and implemented the models in Nvidia Modulus\footnote{https://developer.nvidia.com/modulus}. RANS-PINN uses a warm-start phase, updating each network independently using their corresponding data loss. Each geometry had an unstructured mesh with automated inflation layers or wake region based refinements. During the sampling of data points for training the RANS-PINN, using uniform sampling can bias the network to shift focus away from the bulk regions with low refinement, which in turn can significantly affect the overall prediction accuracy. We adopt a zone-based sampling approach to address this issue and limit the maximum number of collocation points sampled from regions within the solution domain (please see \ref{Sampling_Detail_Appdx} for additional details). Following the warm-start phase, we introduce the PDE constraints into the loss function. Moreover, to normalize the effect of the individual components of the PDE loss function, we scale them by the inverse of their corresponding residual values. We then use \textit{Adam} with a decaying step size (with the initial step size of 0.001 and a decay rate of 0.95) until the training loss converges. For each of the individual output variables (i.e., $u$, $v$, $p$, $k$, and $\epsilon$), we use separate neural networks, all sharing the same input variables consisting of positional coordinates $(x,y)$ and the associated Reynolds number. These networks are connected to the supervised/data loss, as well as the nodes of the PDE loss components. 
\section{Results and Discussions}
\label{headings}
\begin{figure}[b!]
\vspace{-1.5em}
$\begin{array}{cc}
\includegraphics[width=0.4825\linewidth]{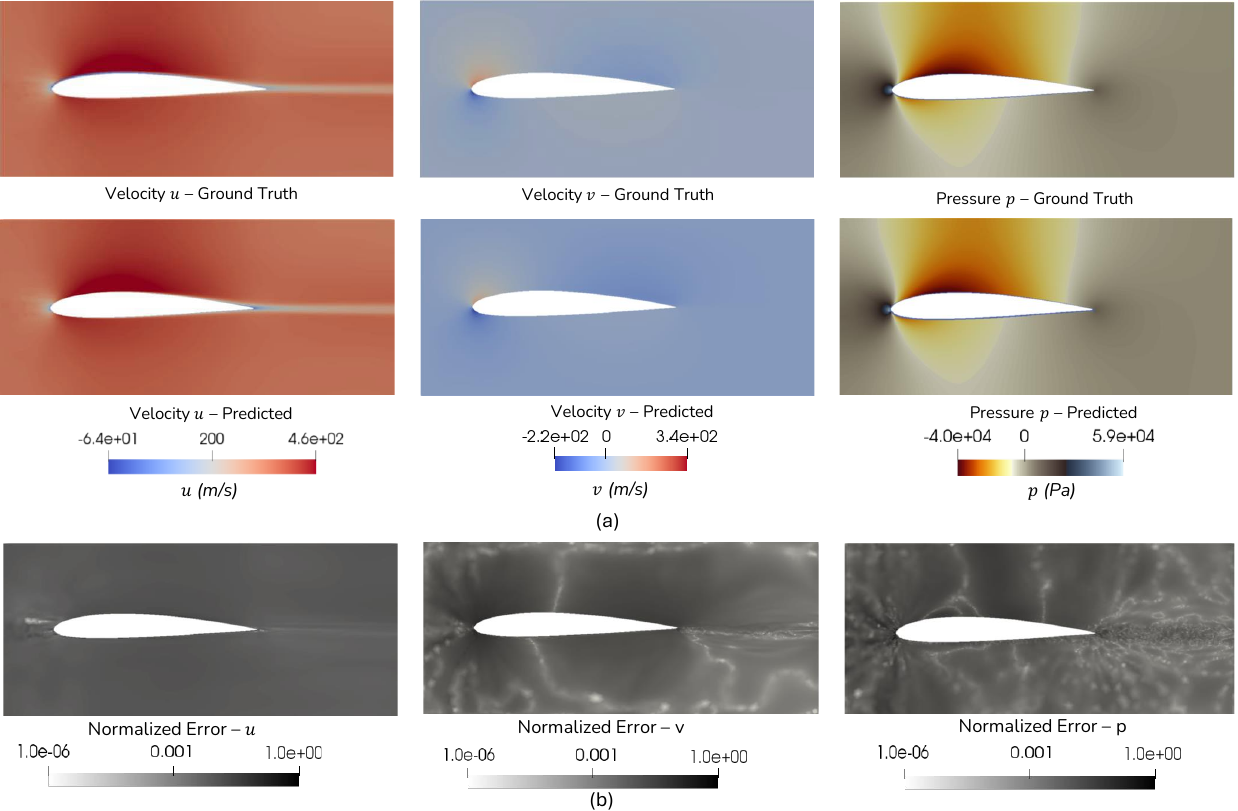}
&  
\includegraphics[width=0.495\linewidth]{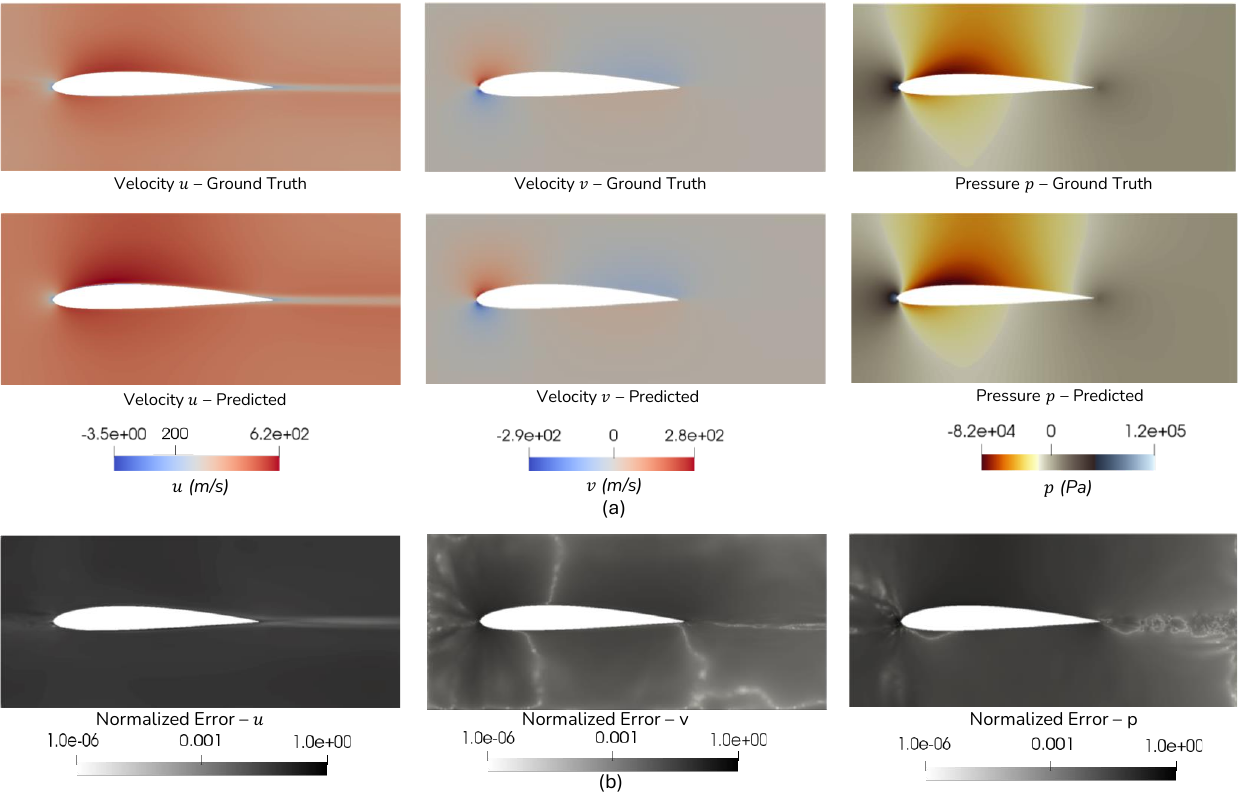}
\\
\small{\textbf{(i)}\;\mathcal{R}_e=375K}
& 
\small{\textbf{(ii)}\;\mathcal{R}_e=460K}
\end{array}$
\vspace{-0.6em}
\caption{\small{(a) \textit{Predicted and true values for primary variables} and the corresponding (b) \textit{relative errors} for flow over an airfoil.}}
\label{airfoil_val}
\end{figure}
\subsection{Airfoil}
As the flow over an airfoil constitutes an external flow, high Reynolds numbers are required for the emergence of turbulence. To learn a parametric surrogate model that can predict flow fields around an airfoil at any given Reynolds number, we generated a training dataset in STAR-CCM+ with six Reynolds numbers ranging from $260K$ to $450K$, spaced equally. This surrogate model plays a critical role in design exploration which involves predicting flow fields over many $\mathcal{R}_e$ values. A tetrahedral unstructured mesh was created with refinements near the surface and in the wake region beyond the airfoil's trailing edge. Similar to the cylinder case discussed by \cite{ghosh2023_RANSpinn}, 3,000 training data points were selected from each flow domain. The boundary conditions included a constant velocity at the inlet, zero pressure at the outlet, free-stream conditions on the top and bottom walls, and no-slip conditions on the airfoil surfaces. Due to its aerodynamic shape, a narrow wake zone forms in the flow field around the airfoil. The surface curvatures of the airfoil cause higher flow acceleration on the top surface, resulting in lower pressures than the bottom surface, which generates lift. This lift is crucial for applications such as turbomachinery and aircraft wings.

Two validation cases were evaluated at Reynolds numbers of $375K$ and $460K$. Figure \ref{airfoil_val}(i) shows the flow field predictions at $\mathcal{R}_e=375K$. The stagnation zone at the airfoil's nose is indicated by a blue region with low $u$-velocity, while high acceleration zones on the top and bottom surfaces are depicted in dark red. The flow separation zone and wake downstream of the airfoil are shown in light blue. The $v$-velocity plots reveal areas of high magnitude with opposing signs near the stagnation zones. The pressure plots show high stagnation pressure at the upstream nose, along with negative pressure zones on both the pressure (bottom) and suction (top) surfaces of the airfoil due to flow acceleration. In the out-of-training-range prediction at $\mathcal{R}_e=460K$ (Figure \ref{airfoil_val}(ii)), a darker shade, indicating a higher $u$-error, is observed due to an overall mismatch in values. The error plots for the two validation cases (bottom (b) panels of Figure \ref{airfoil_val}) reveal very minute dark regions of high magnitude at the stagnation zones for both velocities and pressure. The thin, aerodynamic shape of the airfoil leads to flow impingement at the upstream stagnation point, followed by a sudden change in the flow direction. These high gradients make it challenging for the PINN to accurately predict magnitudes in these areas. Please see \ref{ErrorAnalFoA} for further details.
\subsection{Backward-facing Step}
The backward-facing step is another well-studied turbulent flow scenario \citep{pitzdbfs} wherein a sudden expansion in the flow channel induces flow separation due to an adverse pressure gradient, leading to the formation of a free shear layer characterized by turbulent mixing. This change creates a steep pressure gradient, diverting the flow outward toward the bottom wall and forming a separation bubble accompanied by a free shear layer. The flow then reattaches to the wall further downstream from the step. We assumed constant inlet velocity boundary condition, zero-pressure outlet, and no-slip walls while simulating this flow in STAR-CCM+ for six Reynolds numbers, ranging from 5,000 to 7,600 and adopted a sampling strategy similar to the airfoil geometry.

\begin{wrapfigure}[22]{r}[-12pt]{0.7\linewidth}
\vspace{-1.20em}
\begin{tabular}{c}
\includegraphics[width=1.02\linewidth]{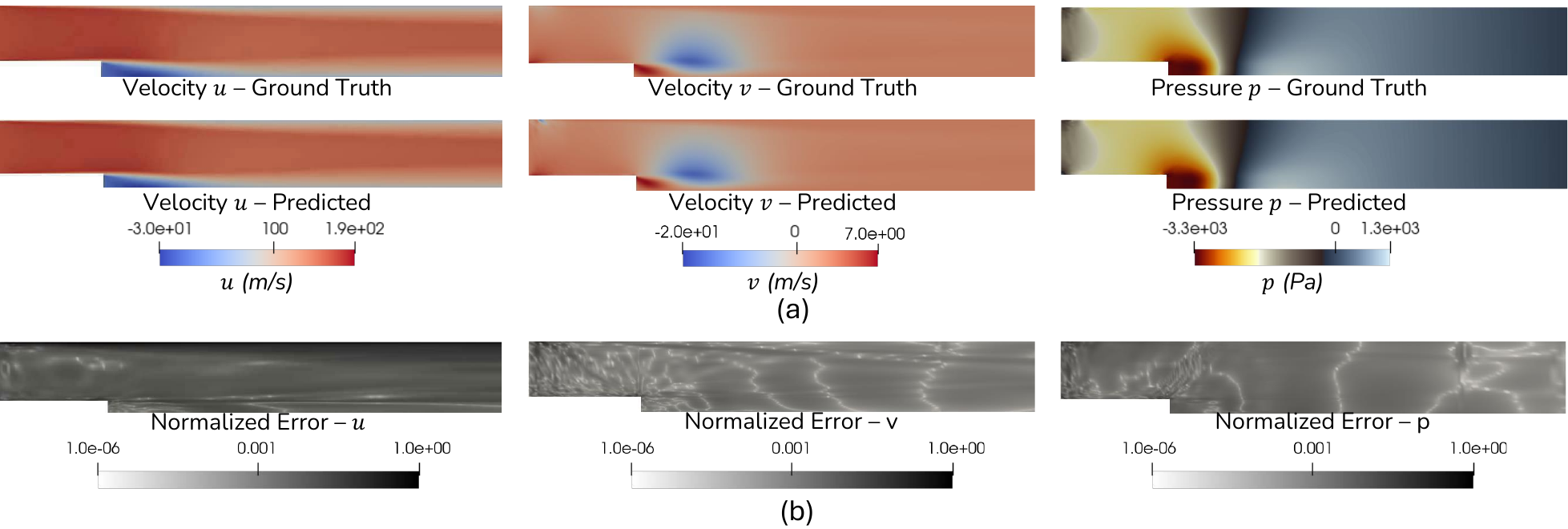}
\\
\small{\textbf{(i)} $\mathcal{R}_e=6400$}
\\
\includegraphics[width=1.02\linewidth]{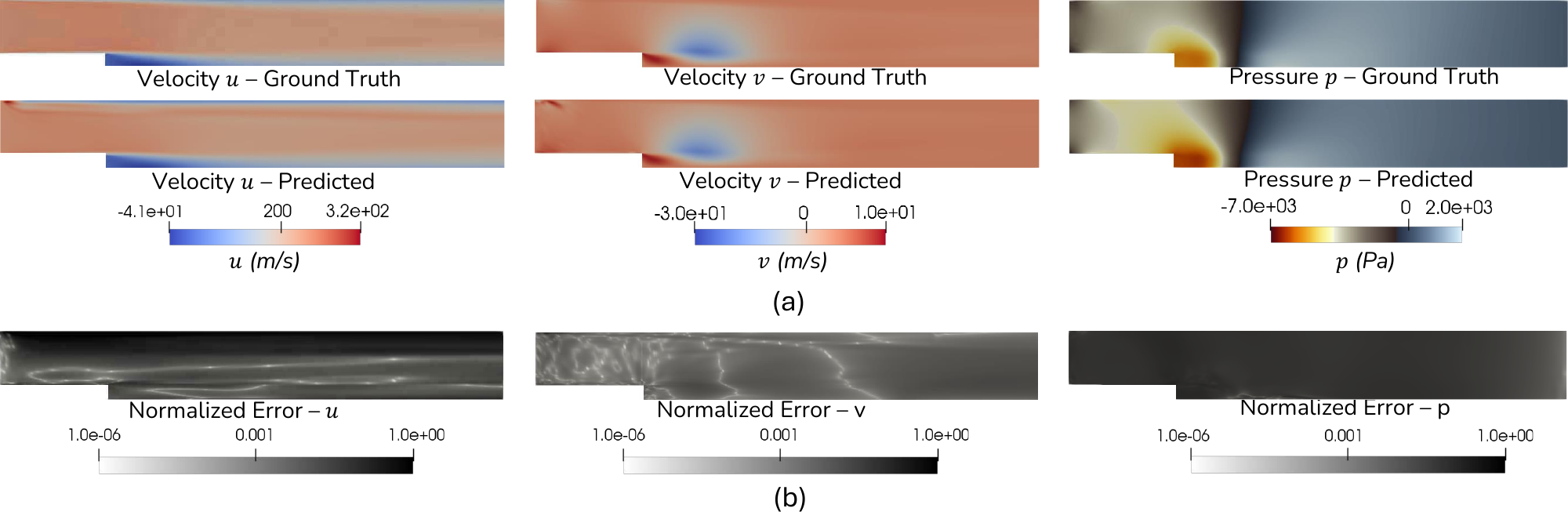}
\\
\small{\textbf{(ii)} $\mathcal{R}_e=8400$}
\end{tabular}
\vspace{-0.8em}
\caption{\small{(a) \textit{Predicted and true values for primary variables} and the corresponding (b) \textit{relative errors} for flow over a backward-facing step.}}
\label{BFS_val}
\end{wrapfigure}

Two validation cases at $\mathcal{R}_e=6400$ (Figure \ref{BFS_val}(i)) and $\mathcal{R}_e=8400$ (Figure \ref{BFS_val}(ii)) were evaluated for this problem. The flow phenomena associated with the backward-facing step can be analyzed from Figure \ref{BFS_val}(i)(a). The $u$-velocity plots reveal a darker region upstream in the channel, indicating higher velocities than the expanded downstream area, which results in a recirculation bubble (seen in blue) near the
step due to the sudden expansion of the flow area. The bulk flow in the expanded zone appears lighter, reflecting lower flow velocity magnitudes. The $v$-velocity plots exhibit more intriguing properties, showing opposing flow directions at the corners of the inlet. These subtleties are well captured by the PINN for the $\mathcal{R}_e=6400$ case but are partially missed at the $\mathcal{R}_e=8400$ case (Figure \ref{BFS_val}(ii)), particularly at the inlet corners. The $u$-velocity boundary layer on the top wall is predicted to be thicker than the ground truth, and a sudden discontinuity is observed in the $v$-velocity at the same location. Given that $\mathcal{R}_e=8400$ is the highest in this study and falls well outside the training range, the model struggles to accurately predict these high gradients. Please see \ref{ErrorAnalBFS} for further details.

\section{Conclusions}
This work investigated the use of parametric PINNs to predict flow fields at unseen Reynolds numbers for incompressible turbulent flows. The RANS formulation, commonly used in industrial applications due to its computational efficiency, can become expensive for complex geometries. The proposed PINN surrogate leverages the two-equation $k-\epsilon$ RANS model to offer a fast alternative, providing accurate predictions crucial for design exploration. The results demonstrate strong predictive capability, particularly in capturing complex turbulent flow features such as separations. While $u$-velocity predictions showed higher normalized errors, reaching up to $12\%$ outside the training range, pressure and $v$-velocity predictions were more accurate, with errors of up to $3\%$ and $2\%$. These findings highlight the potential of PINNs for fast and reliable flow predictions in complex, turbulent scenarios.
\newpage
\small
\bibliography{MainPaper.bib}
\bibliographystyle{abbrvnat}

\newpage
\small
\appendix

\section{Supplemental Material}
\subsection{Overall architecture}
\begin{figure}[H]
\centering

\includegraphics[width=0.72\textwidth]{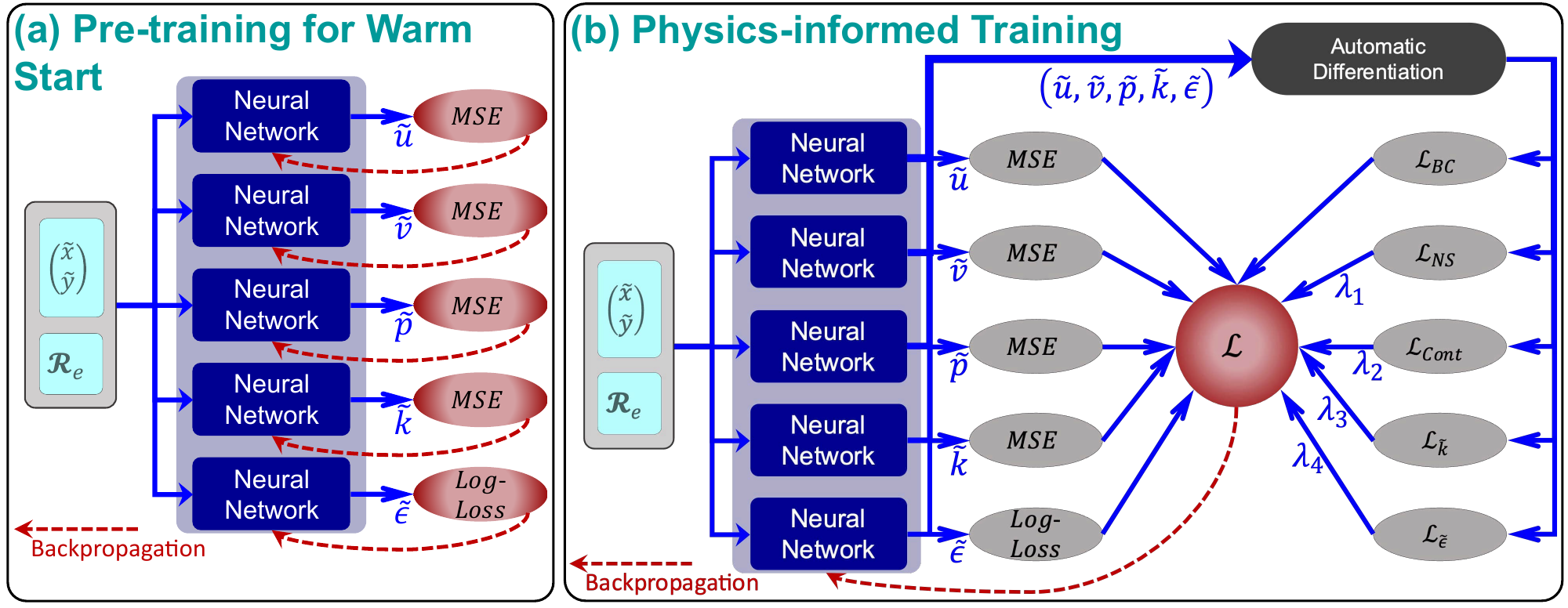}
\caption{\small{Overall RANS-PINN framework for learning surrogates to predict turbulent flow. Left panel shows the warm-up stage, with data losses. Right panel shows the second stage with PDE+data losses and individual PDE loss weights.}}
\label{arch_PINN1}
\vspace{-1.5em}
\end{figure}

The RANS-PINN architecture (Fig.~\ref{arch_PINN1}) uses neural networks to predict variables of interest. These neural networks can be realized via MLPs, FNOs, or even KANs. In our experiments, we have used Fourier neural operators\citep{li2021fourier} (with their default hyperparameters used in Nvidia Modulus) for the cases of cylinder and airfoil, and feed forward networks for the flow over a backwards facing step. For each of the individual output variables (i.e., $u$, $v$, $p$, $k$, and $\epsilon$), we use separate neural networks, all sharing the same input variables consisting of positional coordinates $(x,y)$ and the associated Reynolds number. These networks are connected to the supervised/data loss, as well as the nodes of the PDE loss components.

\begin{figure*}[t!]
\centering
\includegraphics[width=0.95\textwidth]{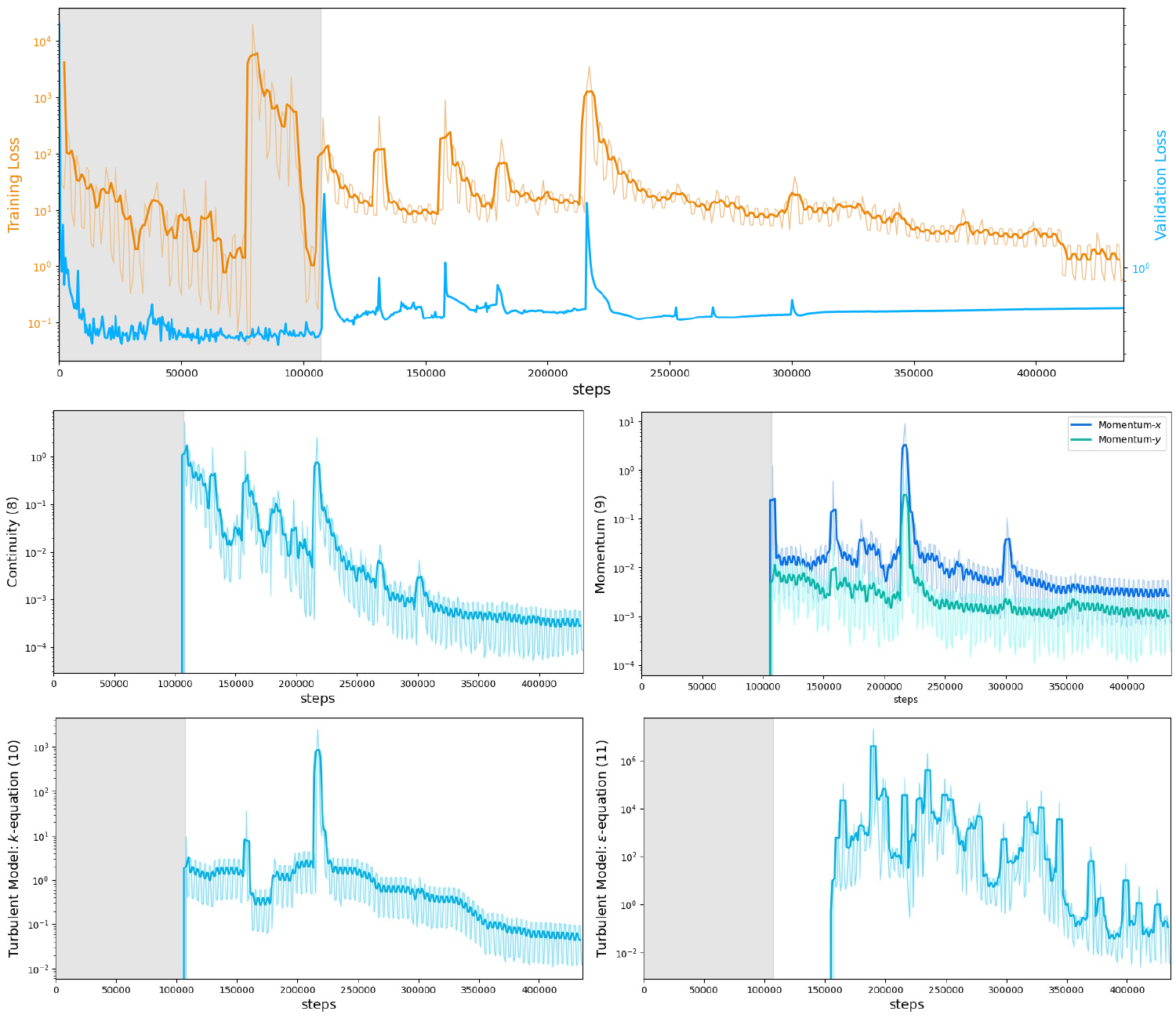}
\caption{\small{Example of the evolution of training losses for PDE components vs epochs. The dark background represents the pre-training stage, where the pde residuals are seen to be 0 in the bottom four plots.}}
\label{loss_evolution}
\vspace{-2em}
\end{figure*}

Conventional approaches to training PINNs involve introducing data and PDE losses simultaneously at the start of the training phase, often with equal weight multipliers. However, they often results in noisy training losses, slow convergence, and high validation error. RANS-PINN addresses these challenges by employing a pre-training step that only uses the data-driven supervised loss. During pre-training, each of the individual networks is updated independently using their corresponding data loss. Following pre-training, we introduce the PDE constraints into the loss function. Moreover, to normalize the effect of the individual components of the PDE loss function, we scale them by the inverse of their corresponding residual values. We then use \textit{Adam} with a decaying step size (with the initial step size of 0.001 and a decay rate of 0.95) until the training loss converges. A sample training case has been shown in Figure \ref{loss_evolution}. It can be seen that the pre-training zone in dark background, has no residual values for the PDE losses. A sudden spike followed by a slow decay is seen at the interface of dark and light backgrounds, indicating the introduction of PDE losses. The epsilon PDE is introduced at a further later stage, as this approach has been found to be beneficial for training convergence. 

To address the challenges associated with abrupt changes observed in the turbulence dissipation term $\epsilon$ near wall and free shear regions, we use a \textit{logarithmic loss function} for both data and PDE losses associated with $\epsilon$. Everything else is computed using an \textit{MSE loss function}. The overall loss function can then be expressed as:
\begin{equation}
\mathcal{L} = \mathcal{L}_{data} + \mathcal{L}_{BC} + \mathcal{L}_{PDE},
\end{equation}
where the PDE loss is defined with weights $\lambda_i$'s as:
\begin{equation}
\mathcal{L}_{PDE} = 
\lambda_1\mathcal{L}_{NS} + \lambda_2\mathcal{L}_{Cont} + \lambda_3\mathcal{L}_{k} + \lambda_4\mathcal{L}_{\epsilon}.
\end{equation}

\subsection{Dataset generation with CFD}
\label{DataGenAppendix}
In this study, we employ Simcenter STAR-CCM+ (\textit{Release 17.02.008} to simulate turbulent flow scenarios using RANS CFD with the $k$-$\epsilon$ turbulence model. Automatic meshers have been used for each case, with refinement near walls for low wall $y+$, and wall functions for turbulence quantities. Moreover, we have used wake refinements to simulate flow around the cylinder and the airfoil. The data generated from the simulation is then normalized using the non-dimensional version of the underlying dynamics (i.e., continuity, Navier-Stokes, and RANS equations). More discussion on the mesh and data sampling can be found in section~\ref{Sampling_Detail_Appdx}.

\subsection{Discussions on Sampling}
\label{Sampling_Detail_Appdx}
\begin{figure}[H]
\centering
\label{ablation_fig}
\includegraphics[width=0.43\textwidth]{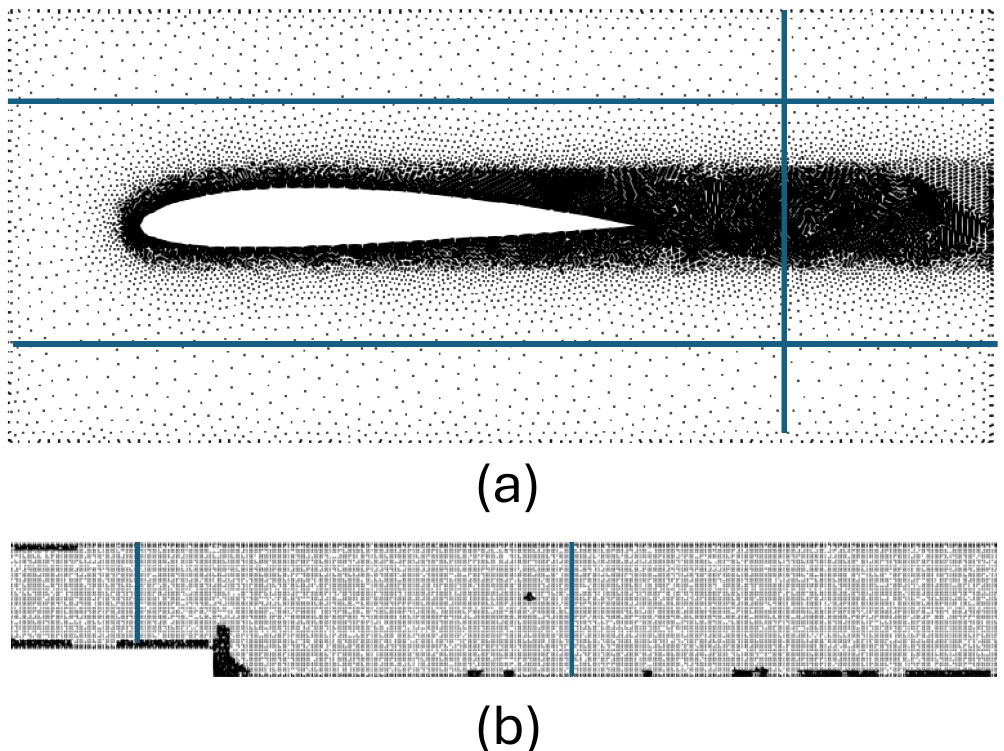}
\caption{\small{Point cloud derived from the full CFD mesh for the three geometries, showing sampling zones.}}
\label{imp_sampl}
\vspace{-1.5em}
\end{figure}

\begin{figure}[H]
\centering
\label{ablation_fig}
\includegraphics[width=0.56\textwidth]{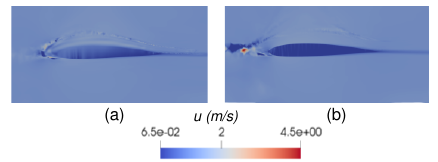}
\caption{\small{Two instances of noisy prediction due to lack of importance sampling.}}
\label{imp_sampl_abl}
\vspace{-1.5em}
\end{figure}

Mesh and sub-regions for the two geometries can be seen in Figure \ref{imp_sampl}, where the sub-divisions in the domain have been created, and then uniform sampling has been carried out. Lack of intelligent sampling has often resulted in poor convergence and noisy predictions. Figure \ref{imp_sampl_abl} shows two instances of noisy prediction from a non-parametric PINN trained on a single $\mathcal{R}_e$ due to a lack of importance sampling. The flow over an airfoil case has refinements near the wake region and the wall, which can result in oversampling in those areas and a lack of sufficient samples from the bulk flow regions away from the airfoil. 
\subsection{Spatial distribution of errors: Flow over airfoil}
\label{ErrorAnalFoA}
The error plots for the two cases (Figure \ref{airfoil_val}) show very minute dark regions of high magnitude at the stagnation zone for all three variables, the velocities, and pressure. The thin, aerodynamic shape of the airfoil results in impingement of flow at the stagnation point upstream, followed by a sudden change in the flow direction. All these high gradients make it difficult for the PINN to predict accurate magnitudes at these locations. A darker shade due to a higher u-error is observed in Figure \ref{airfoil_val} (ii) due to an overall mismatch in values. The variance plots are again dominated by low values below $10^{-3}$, and no discernible pattern is recognized across the cases. Table\ref{error_airfoil} shows the numerical error metrics for the 2 Reynolds numbers. The highest mean error in all cases is seen for the u-velocity, with the maximum error of 12$\%$ in the case of $\mathcal{R}_e=460K$. The rest of the values show less than 1$\%$ error.
\begin{table}[H]
  \centering
  \renewcommand{\arraystretch}{1.2}
  \begin{tabular}{|p{1cm}|c|c|c|c|c|c|c|c|c|}
    \hline
    \multirow{2}{2cm}{$\mathcal{R}_e$} & \multicolumn{3}{c|}{\textbf{Normalized $u$-error}} & \multicolumn{3}{c|}{\textbf{Normalized $v$-error}} & \multicolumn{3}{c|}{\textbf{Normalized $p$-error}}\\
    \cline{2-10}
    & \textbf{Mean} & \textbf{95th p} & \textbf{Median} & \textbf{Mean} & \textbf{95th p} & \textbf{Median}&
    \textbf{Mean} & \textbf{95th p} & \textbf{Median} \\
    \hline
    375K & 
0.06 & 0.094 & 0.054 & 0.005 & 0.017 & 0.001 & 0.002 & 0.008 & 0.0005 \\ \hline   
460K & 0.12 & 0.1900 & 0.025 & 0.0013&0.04&0.0007&0.031 &0.059&0.031\\ \hline
\end{tabular}
\vspace{0.5em}
\caption{\small{Error Metrics for flow over a NACA 2412 airfoil.}}
\label{error_airfoil}
\end{table}
\begin{figure*}[h]
\vspace{-2.5em}
\centering
\includegraphics[width=0.85\textwidth]{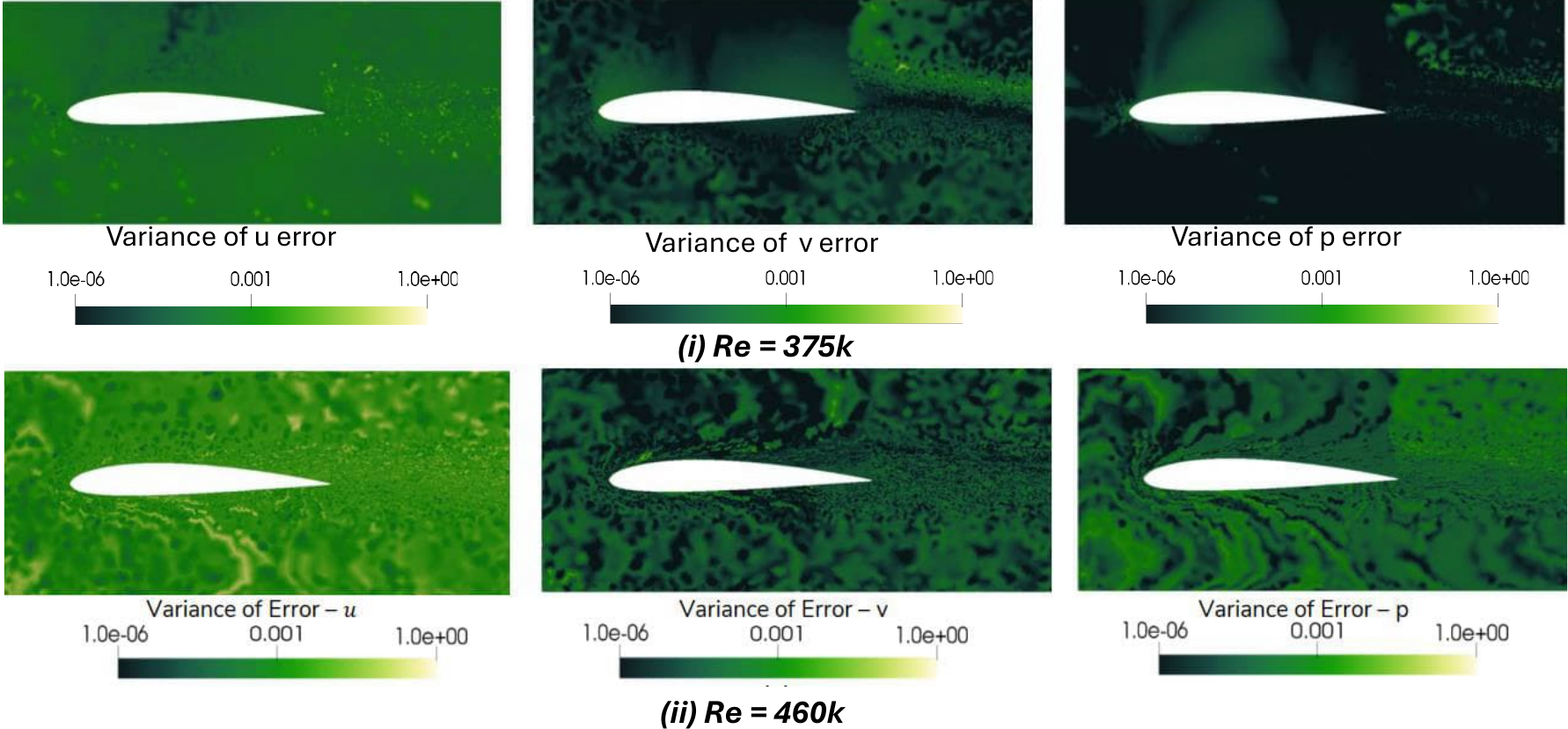}
\caption{\small{Variance of errors for flow over an airfoil, based on repeated random initializations of weights and biases}}
\label{arch_PINN}
\vspace{-0.5em}
\end{figure*}

\begin{figure}[h!]
\vspace{-1.5em}
	\centering
	\begin{subfigure}{}
		\includegraphics[width=\textwidth]{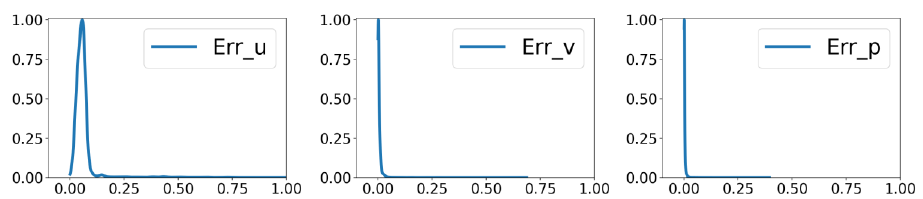}
		\small{\textbf{(a)} }
	\end{subfigure}
	\begin{subfigure}{}
		\includegraphics[width=\textwidth]{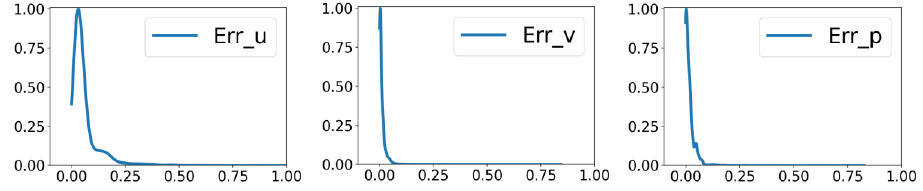}
		\small{\textbf{(b)} }
	\end{subfigure}
	\caption{\small{Validation error histogram of $x$-velocity, $y$-velocity and pressure for both validation cases (a) $\mathcal{R}_e=375K$ and (b) $\mathcal{R}_e=460K$ of the flow over a NACA 2412 airfoil.}}
        \label{error_hist_airfoil}
        \vspace{-1em}
\end{figure}

It is, however, again important to understand the distribution of error over the population of the spatial points, which has been plotted in Figure\ref{error_hist_airfoil}. A slight hint of a bimodal nature of the error vs the fraction of population is seen in the (a)  case for $\mathcal{R}_e=375K$. The bimodal peak is more distinct in the case of the out-of-range $\mathcal{R}_e=460K$, especially for the $u$-velocity and pressure, which explains the higher mean error. Despite this, the median error is low for this case due to the overall skewness of the distribution towards zero, indicating the higher presence of points with low error. Overall, all the variables had a skewed curve with peaks very close to the zero mark. This resulted in high disparities between the 95th percentile values and mean/median values for the two values of $\mathcal{R}_e$, especially the $\mathcal{R}_e=375K$ case in Table\ref{error_airfoil}. 

\subsection{Spatial distribution of errors: Backwartd-facing step}
\label{ErrorAnalBFS}
\begin{figure}[h!]
	\centering
	\begin{subfigure}{}
		\includegraphics[width=\textwidth]{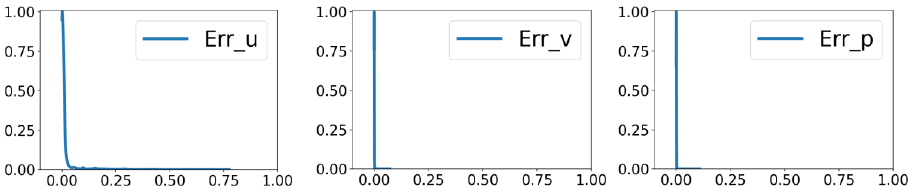}
	\small{\textbf{(a)} }	
	\end{subfigure}
	
	\begin{subfigure}{}
		\includegraphics[width=\textwidth]{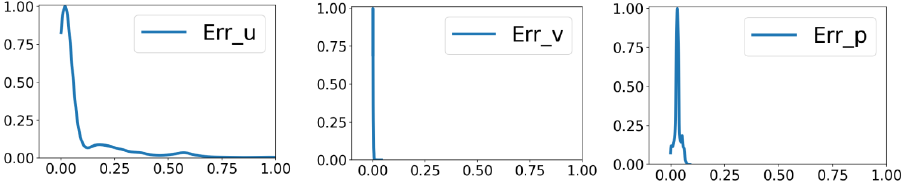}
	\small{\textbf{(b)} }	
	\end{subfigure}
	\caption{\small{Validation error histogram of $x$-velocity, $y$-velocity and pressure for both validation cases (a) $\mathcal{R}_e=6400$ and (b) $\mathcal{R}_e=8400$ for the backward-facing step.}}
       \label{BFS_histogram}
\end{figure}

The pressure plots in Figure \ref{BFS_val} reflect the sudden changes due to velocity profile development (at the inlet), where the positive pressure drops in magnitude in the downstream direction till the step region, where a sudden increase in pressure happens after the flow area expansion. Due to the sharp nature of the step geometry, the gradients created by the pressure changes are very sharp. Also, due to flow reattachment, the pressure sign changes from negative to positive downstream of the step and then follows a decreasing trend till it reaches the zero magnitude at the exit. The normalized error plots (Figure \ref{BFS_val}) display differences between predicted and ground truth flow variables in grayscale. Narrow dark regions are then observed near the upper walls for the $u$-velocity errors, signifying a mismatch in the boundary layer formation prediction. A small zone of dark patch is noticed in the top left corner, where the flow development occurs. The pressure plots also have high errors in co-incidental spots. For the $\mathcal{R}_e=8400$ validation case (Figure \ref{BFS_val})(ii), the high error bands are particularly larger due to higher mismatch, but the locations of those zones are the same as the other two cases. This observation is reflected in Table \ref{error_BFS}, where the highest mean and median errors are seen for all three variables in the $\mathcal{R}_e=8400$ case. The other case shows the highest mean errors for the $u$-velocity, with max value at 1$\%$ for $u$, and lower than $1\%$ for the other flow variables. 
\begin{table}[t!]
  \centering
  \renewcommand{\arraystretch}{1.2}
  \begin{tabular}{|p{1cm}|c|c|c|c|c|c|c|c|c|}
    \hline
    \multirow{2}{2cm}{$\mathcal{R}_e$} & \multicolumn{3}{c|}{\textbf{Normalized $u$-error}} & \multicolumn{3}{c|}{\textbf{Normalized $v$-error}} & \multicolumn{3}{c|}{\textbf{Normalized $p$-error}}\\
    \cline{2-10}
    & \textbf{Mean} & \textbf{95th p} & \textbf{Median} & \textbf{Mean} & \textbf{95th p} & \textbf{Median}&
    \textbf{Mean} & \textbf{95th p} & \textbf{Median} \\
    \hline
    6400&0.015 &0.096&0.002&0.0003&0.009&0.0001&0.0007&0.0018&0.0005
    \\\hline
    8400&0.12&0.585&0.025& 0.023& 0.13&0.002 &0.031&0.057&0.031
    \\ \hline
  \end{tabular}
  \vspace{0.5em}
  \caption{\small{Error Metrics for flow over a backward-facing step.}}
  \label{error_BFS}
  \vspace{-2em}
\end{table}
%
%
\begin{figure}[h]
\vspace{-0.5em}
\centering
\includegraphics[width=0.85\textwidth]{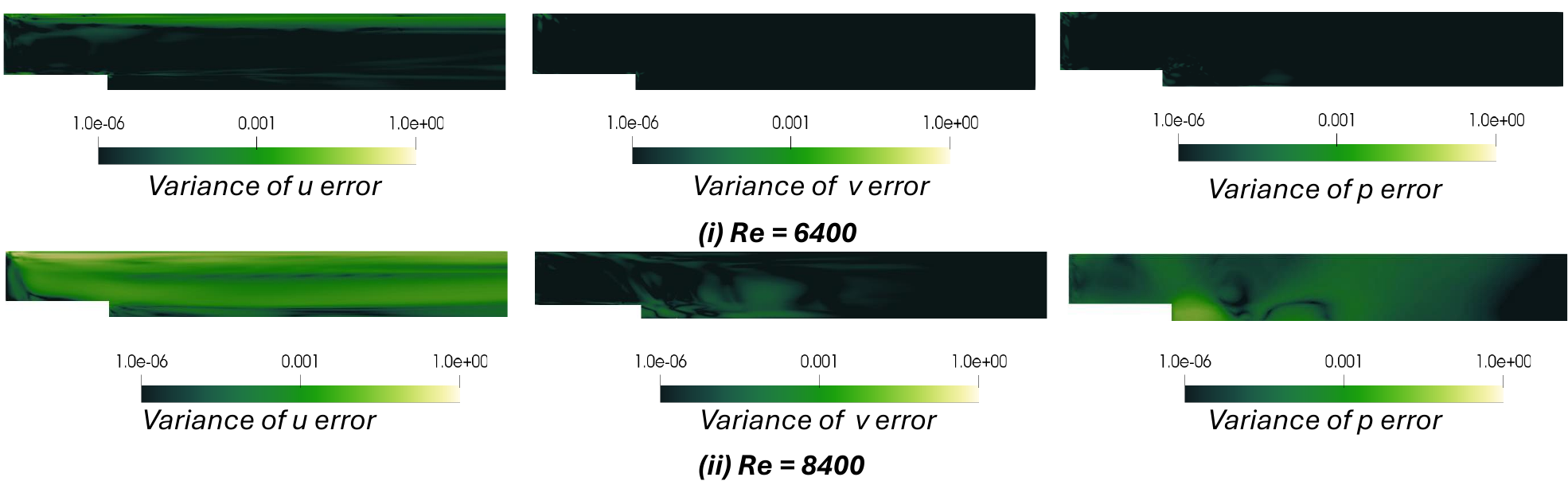}
\vspace{-0.25em}
\caption{\small{Variance of errors for Backward-facing step, based on repeated random initializations of weights and biases}}
\label{arch_PINN}
\vspace{-0.75em}
\end{figure}
The lower values of the median compared to the mean suggest a higher population of data points (spatially distributed) has low error metrics. Thus, it is important to look at the spatial distribution of errors and also the population distribution of error magnitudes. This has been done in Figure \ref{BFS_histogram}, where the peaks of all the distributions are seen quite close to the 0 error mark, and the distributions themselves are highly skewed towards the $y$-axis, except the $\mathcal{R}_e=8400$ case. Here, a slightly higher width is noticed for the peak, with a shift towards the right. Two smaller additional peaks are also seen for the $u$-velocity case, which hints towards a trimodal nature of the distribution. This indicates a higher population of high error magnitude points. Similarly, the $p$ variable has a smaller additional peak.

\end{document}